\documentclass[review]{elsarticle}

\usepackage{lineno,hyperref}
\usepackage{tabularx}
\usepackage{tabulary}

\usepackage{array}
\newcolumntype{P}[1]{>{\centering\arraybackslash}p{#1}}
\usepackage{multirow}
\usepackage{graphicx}
\usepackage{subcaption}
\usepackage{rotating}
\usepackage{lipsum}
\usepackage{lscape}
\usepackage{amsmath,amssymb,amsthm,mathtools}
\usepackage{lmodern}

\usepackage[table,xcdraw]{xcolor}
\usepackage[linesnumbered,ruled,vlined]{algorithm2e}
\usepackage{textcomp}
\usepackage{float}
\usepackage[margin=1in]{geometry}
\modulolinenumbers[5]
\usepackage{pdflscape}
\journal{Expert Systems with Applications}

\DeclarePairedDelimiter\floor{\lfloor}{\rfloor}

\usepackage{numcompress}\biboptions{authoryear}

\begin{document}

\begin{frontmatter}
\title{A deep learning approach to predict the number of $k$-barriers for intrusion detection over a circular region using wireless sensor networks }

\author[iiserb]{Abhilash Singh}
\author[gbu]{J. Amutha}
\author[iitk]{Jaiprakash Nagar}
\author[mits]{Sandeep Sharma \corref{san}}
\cortext[san]{Corresponding author\\ Email address: sandeepsvce@gmail.com (Sandeep Sharma) ORCID: 0000-0002-1098-7633} 

\address[iiserb]{Fluvial Geomorphology and Remote Sensing Laboratory, Indian Institute of Science Education and Research Bhopal, India}
\address[gbu]{University School of ICT, Gautam Buddha University, Greater Noida, India}
\address[iitk]{Subir Chowdhury School of Quality and Reliability, Indian Institute of Technology Kharagpur, India}
\address[mits]{Madhav Institute of Technology and Science, Gwalior, Madhya Pradesh, India


}




\begin{abstract}
Wireless Sensor Networks (WSNs) is a promising technology with enormous applications in almost every walk of life. One of the crucial applications of WSNs is intrusion detection and surveillance at the border areas and in the defense establishments. The border areas are stretched in hundreds to thousands of miles, hence, it is not possible to patrol the entire border region. As a result, an enemy may enter from any point absence of surveillance and cause the loss of lives or destroy the military establishments. WSNs can be a feasible solution for the problem of intrusion detection and surveillance at the border areas. Detection of an enemy at the border areas and nearby critical areas such as military cantonments is a time-sensitive task as a delay of few seconds may have disastrous consequences. Therefore, it becomes imperative to design systems that are able to identify and detect the enemy as soon as it comes in the range of the deployed system. In this paper, we have proposed a deep learning architecture based on a fully connected feed-forward Artificial Neural Network (ANN) for the accurate prediction of the number of $k$-barriers for fast intrusion detection and prevention. We have trained and evaluated the feed-forward ANN model using four potential features, namely area of the circular region, sensing range of sensors, the transmission range of sensors, and the number of sensor for Gaussian and uniform sensor distribution. These features are extracted through Monte Carlo simulation. In doing so, we found that the model accurately predicts the number of $k$-barriers for both Gaussian and uniform sensor distribution with correlation coefficient (R = 0.78) and Root Mean Square Error (RMSE = 41.15) for the former and R = 0.79 and RMSE = 48.36 for the latter. Further, the proposed approach outperforms the other benchmark algorithms in terms of accuracy and computational time complexity.
\end{abstract}


\end{frontmatter}

\section{Introduction}

The unquenchable thirst of people for political and military power is compelling them to extend their boundaries and grab other people's natural resources. In order to achieve this goal, they may try different techniques such as gaining information about the military establishments, the number of military personnel at a given place, regions of natural resources, and the vulnerabilities of authorities that they can exploit. In addition, illegal immigration, smuggling of drugs, and other banned commodities across the boundaries are immediate concerns that must be dealt with immediately. Therefore, it is crucial to identify an intruder or an unauthorised activity accurately in a timely manner as they all are time-sensitive issues that may result in havoc if not prevented in time. Furthermore, encroachment in the border areas and unauthorised entry in the prohibited regions is a serious issue making border surveillance mandatory.



It is a well-known fact that most real-life problems can be solved with the help of suitable technologies. Fortunately, WSNs are widely used and is a popular technology that can resolve the concerned issue~\citep{keung2012intrusion,huang2018intrusion,singh2021nature,singh2019mathematical}. WSNs are widely deployed for various military applications such as intrusion detection in border areas, combat monitoring, an unauthorised access to prohibited areas, land mines detection, battlefield surveillance, reconnaissance, and so on~\citep{si2020energy,sharma2020intrusion,kandris2020applications,amutha2020wsn}. 

Researchers have proposed various border surveillance and intrusion detection techniques using WSNs \citep{mostafaei2018border, amutha2021distributed, benahmed2019optimal,aseeri2017detection,karthick2019internet,karanja2021development, arjun2019panchendriya,gavel2021maximum}. The proposed techniques use either simulations or Internet of Things (IoT) methods to validate their techniques. However, simulation methods for the validation of proposed techniques have high time complexity, \textit{i.e.}, the time taken to obtain a single output at a given value of sensor and sensing range is in several hours. Also, IoT devices are very expensive and require a huge amount of financial investment. The high time complexity and financial issues can be minimised to a negligible level using machine learning approaches to validate and predict the performance of WSNs before their actual deployment in a given region \citep{mishra2018detailed,singh2021gaussian,kotiyal2021ecs}. However, the accurate and timely detection and prevention of intrusion through machine learning approaches is still an ill-posed problem that has been insufficiently investigated. To address this issue, we propose a deep learning architecture for accurate and timely intrusion detection and prevention.

In this paper, we proposed a fully connected feed-forward ANN architecture for the prediction of the number of $k$-barriers for accurate intrusion detection and prevention in WSNs using potential features. We have extracted four features, namely area of the circular region, sensing range of sensors, the transmission range of sensors, and the number of sensors through the Monte Carlo simulation approach. Afterward, we used these features to trained and evaluated the performance of the feed-forward ANN model using R, RMSE, bias, and computational time complexity as performance metrics.   

Further, the rest of the paper is divided into seven sections. In Section \ref{sec:rw}, we have discussed the related works. In Section \ref{sec:sys-model}, we have presented the system model. In this section, we have discussed the sensor distribution models and the sensing model. In Section \ref{sec:sim-exp}, we have discussed the simulation experiment. In Section \ref{sec:mlmodel}, we have discussed the machine learning model. In this section, we have discussed the feature importance, feature sensitivity, and model setup. In Section \ref{sec:results}, we have presented the results. Lastly, in Section \ref{sec:discussion} and \ref{sec:conclusion}, we have presented the discussion and conclusion of this study, respectively.



\section{Related Works}
\label{sec:rw}

Deep learning is a subset of machine learning algorithms which has been applied for intrusion detection using WSNs \citep{lee2021towards,amutha2021strategies,singh2022automl,sood2022intrusion,singh2022lt}. It is also employed for pattern matching and network security where it identifies the malicious activities occurring in the network and is termed as Network Intrusion Detection System (NIDS). The accuracy of the NIDS can be improved with the help of Recurrent Neural Network IDS (RNNIDS) \citep{sohi2021rnnids} which is capable to identify the complex patterns resulting in an enhanced intrusion detection rate. Authors in \citet{yin2017deep} have proposed an RNNs approach which examines the system behaviour, type of intrusion, and the impact of intrusion on the intrusion detection accuracy with the help of learning rate and the number of neurons. The major limitation of RNNIDS is that it fails to minimise the false positives; thus, not able to achieve the maximum detection rate. This issue has been dealt with the help of a deep learning architecture that combines classifiers with Convolution Neural Networks (CNN) having Long Short Term Memory (LSTM), thus, offering maximum detection rate \citep{pektacs2019deep}. Here, the CNN acquires spatial information, and the LSTM acquires temporal features from the received packets in the network. The optimal parameters in the feature space are obtained by employing an estimator known as tree-structured Parzen. This method focuses on generating flow-based statistical features rather than data-set features. Although the abnormal traffic can be detected with an accuracy rate of 99.09 \% and false alarm rate 0.0227, it fails to compute the computational time complexity of flow-based intrusion detection method. 

A deep learning approach that uses an auto-encoder strategy exhibits an improvement in the time complexity by 18.12\% \citep{abbasi2021deep}. This strategy uses multi-layer perception to replace the detection of hierarchical features and unsupervised feature learning. Thus, it enhances the pattern matching mechanism for intrusion detection with the help of Deep Learning-based Feature Extraction (DLFE), which is an Optimisation of Pattern Matching (OPM) approaches. Another approach called Scale-Hybrid-IDS-AlertNet (SHIA) \citep{vinayakumar2019deep} performs data computation at the network and the host-level for the intrusion detection and to deliver the relevant alert notifications to the controller automatically. A Deep Neural Network (DNN) can render an effective IDS that can detect and classify the intruders crossing the Region of Interest (RoI). Here, the performance of SHIA is measured in terms of multi-class classification of the DNN, accuracy, True Positive Rate (TPR), and False-Positive Rate (FPR). Although SHIA is scalable and shows improved performance in handling large data-sets of real-time systems, it does not compute the number of $k$-barriers and incurs high computational cost.

Despite having a high computational complexity, ensemble-based techniques have a high level of accuracy as compared to the base models. An ensemble based DNN framework for the continuous analysis of intrusion detection along with the ability to learn hierarchical data-sets automatically has been proposed in \citet{folino2021learning}. Here, a log-stream of an intrusion detection system maintains an ensemble that contains classifiers trained on discrete chunks of the data-set instance and a combiner model. The combiner model performs reasoning on the instance parameters and classifier predictions; after that, the efficacy is computed in terms of data scarcity and accuracy that allows to analyse the divers ensemble aggregation strategies. This ensemble approach utilises the unstructured data that does not comply with the transfer learning strategies and is found in the log-stream of NIDS.

Barrier coverage using WSNs plays a crucial role for the detection of an unauthorised personnel or object trying to enter the prohibited region. Barriers are formed by the sensors deployed over the entire given RoI. The early depletion of sensor's energy leads to the failure of the sensors causing blind spots in the barriers. To overcome this issue, authors in \citet{saraereh2021robust} have proposed a set-based max-flow procedure for the re-deployment of mobile sensors. This method determines the vulnerable locations and deploys the mobile sensors which strengthens and prolongs the longevity of the barrier. Although the algorithm exhibits higher efficiency in terms of the computation time, it fails to predict the number of $k$-barriers for the IDS. In \citet{nagar2018k}, the authors have derived an analytical closed-form expression using mobile sensors for the $k$-barrier coverage probability of a WSN. Here, they have calculated the total area covered by an intruder traveling at a given angle to cross the RoI, Then, this total area is utilised to obtain the closed-form expression for the $k$-barrier coverage probability of the WSN. In an another work presented in \citet{singh2021gaussian}, the authors have employed three machine learning approaches, namely Gaussian Process Regression (GPR), Scaling GPR (S-GPR), and Center mean GPR (C-GPR) to predict the $k$-barrier coverage probability of a WSN. The proposed GPR technique quickly detects and prevents any intrusion taking place at any location in the RoI. A three-level hierarchy scheme to detect a mobile intruder in distributed WSNs is proposed in \citet{sharma2020sensor} improving the precision of intrusion detection. To maximise the probability of intrusion detection, the authors have used divers sensing techniques, $k$-mean clustering, and the Likelihood Ratio Test (LRT) methods. The LRT fusion rule used in this scheme is efficient in terms of metrics like detection probability and false alarm rate at a given number of sensors and the speed of the mobile intruder. 

The algorithms and strategies proposed in the literature fail to accurately predict the number of $k$-barriers for the intrusion detection. Hence, the overall aim of this study is to overcome the limitation of the previous studies in terms of accuracy and computational time complexity using a deep learning approach.

\section{System Model}
\label{sec:sys-model}

This section briefly discusses sensor distribution models, sensing range model, and the performance metric, namely the number of $k$-barriers. 

\subsection{Sensor Distribution Model}

The choice of sensor distribution model depends on the required application. In this work, we consider two-sensor distribution models; (i) Gaussian distribution model and (ii) uniformly distribution model.

\begin{figure}[t]
    \centering
    \includegraphics[width=0.75\textwidth]{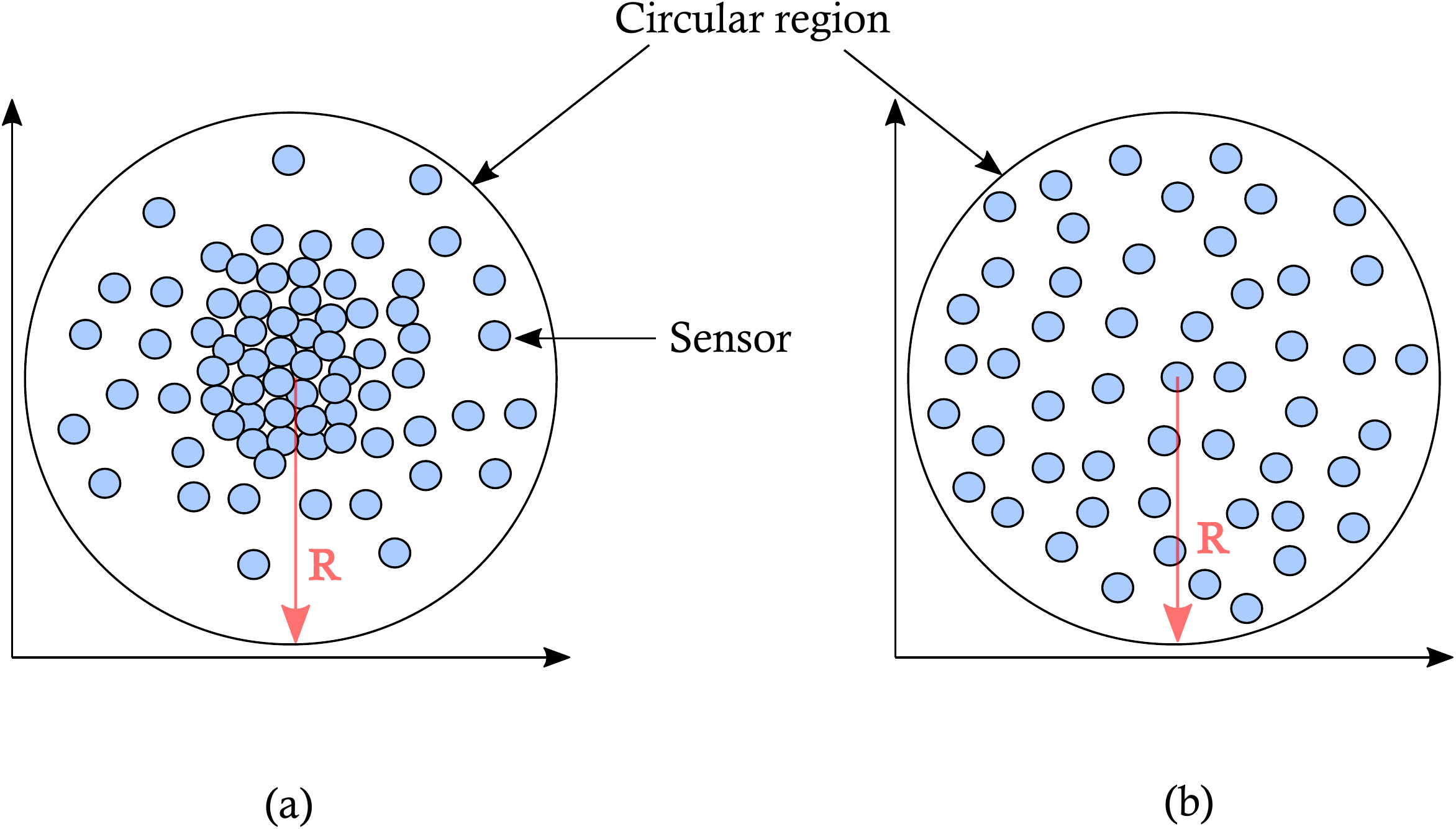}
    \caption{Illustration of (a) Gaussian sensor distribution and (b) Uniform sensor distribution. The blue-filled circles represent sensors.}
    \label{fig:deployment}
\end{figure}

\subsubsection{Gaussian Sensor Distribution Model}

In this model, a finite number of sensors are installed in a finite circular region of radius $R$ meters following a 2D Gaussian distribution, also known as a normal distribution (Fig. \ref{fig:deployment}a). Thus, the Probability Density Function (PDF) for a location $(x,y)$ to be installed with a sensor is given by~\citet{wang2008coverage}

\begin{equation}
\label {eqn:Gaussian distributed SNs 1}
f(x, y) = \frac{1} {2\pi\sigma_x\sigma_y}e^{-(\frac{(x-x_c)^2}{2\sigma_x^2} + \frac{(y-y_c)^2}{2\sigma_y^2})}
\end{equation}

where, $(x_c, y_c)$ represents the centre of the circular region, $\sigma_x$ and $\sigma_y$ are the standard deviations of $x$ and $y$ location coordinates respectively. Furthermore, the location of a sensor inside the circular region represented by $(\gamma, \phi)$ can also be modeled with the help of position coordinates $(x,y)$ if 

\begin{equation}
\label {eqn:Gaussian distributed SNs 2}
f(x, y):\sqrt{(x-x_c)^2 + (y-y_c)^2} \leq R = f(\gamma, \phi):\gamma \leq R
\end{equation}

\subsubsection{Uniform Sensor Distribution Model}

In this model, a finite number of sensors are distributed uniformly and randomly inside a finite circular region (denoted by $\Re$) of radius $R$ meters (Fig. \ref{fig:deployment}b). The position of a random sensor within the circular region is represented by $P_c = (\gamma, \phi)$, where, $\gamma \in [0, R]$, denotes the distance of the sensor from the center of the circular region, and $\phi \in [0, 2\pi]$, denotes the angle between the x-axis and the line that passes through the sensor location. The resulting sensor distribution probability density function is given by

\begin{equation}
\label {eqn:uniformly distributed SNs 1}
f_p(\Re) =
\begin{cases}
 1, if P_{c} \in \Re \\ 
 0, otherwise
 \end{cases}
\end{equation}

Furthermore, the probability that a sensor is located at an arbitrary position $P_c = (\gamma, \phi)$ inside the circular region is given by

\begin{equation}
\label {eqn:uniformly distributed SNs 2}
f(P_c) = \frac{1} {\pi R^2 }
\end{equation}

\subsection{Binary Sensing Range Model}

The binary sensing range model is one of the most widely employed sensing range models for estimating the performance of WSNs \citep{laranjeira2014border, nagar2020analytical}. A point $i$ denoted by $P_i(x_i, y_i)$ inside the 2D circular region will be covered by a sensor $j$ located at $S_j(x_j, y_j)$, if the Euclidean distance of point $i$ from the sensor $j$ is less than or equal to the sensing range $r_s$ of the sensor. Mathematically, it can be represented as 

\begin{equation}
\label {eqn:binary}
P(S_i)=
\begin{cases}
1, if d(S_i,P_i) \leq r\textsubscript{s}  \\
 0, otherwise
\end{cases}
\end{equation}

\subsection{Coverage Graph}

\begin{figure}
    \centering
    \includegraphics[width=0.75\textwidth]{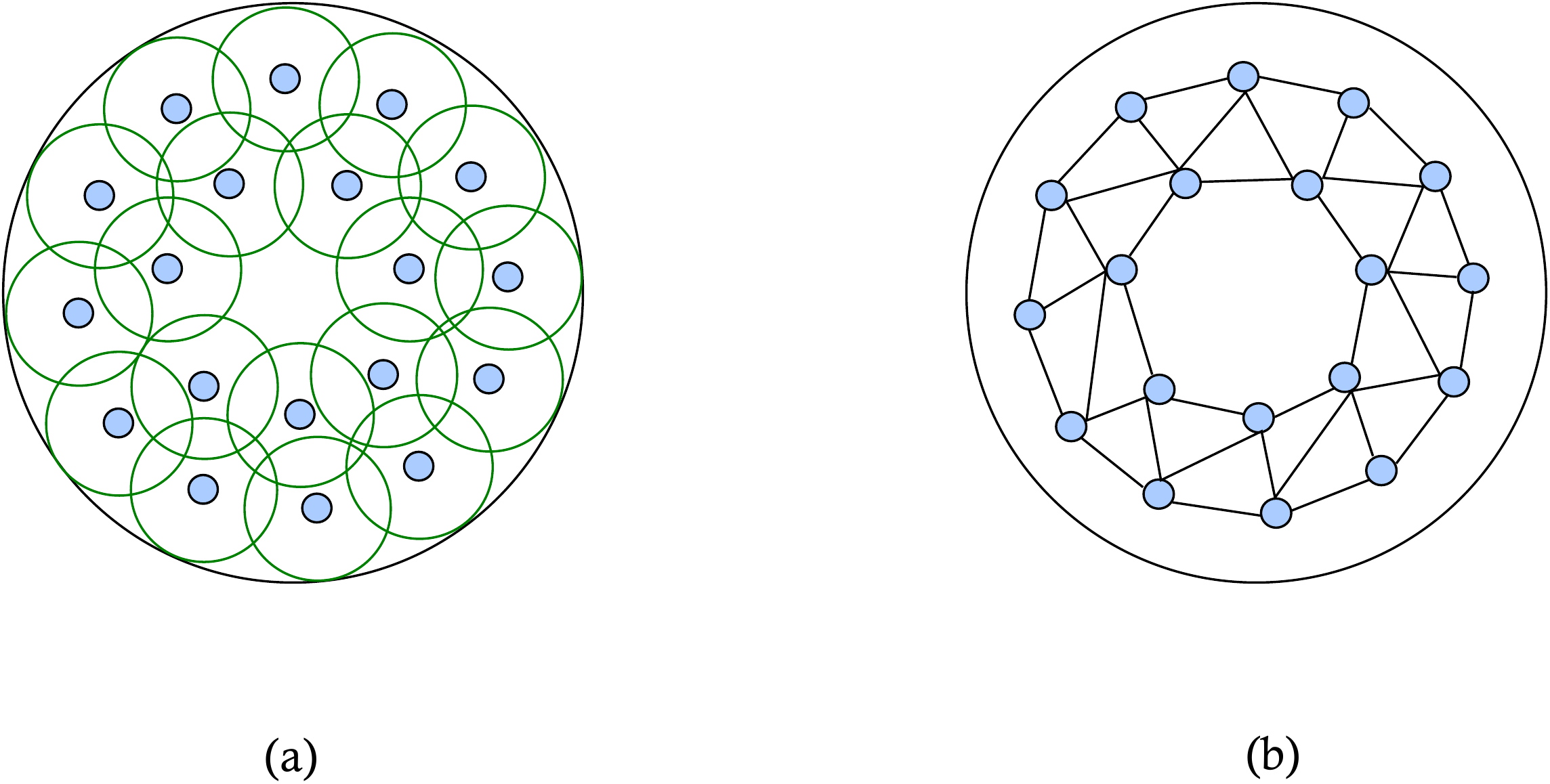}
    \caption{Illustration of (a) 2-Barrier coverage and (b) Coverage graph.}
    \label{fig:cir}
\end{figure}

Coverage is the measure of how well the sensors are monitoring the RoI in which they are deployed. The $k$-coverage ensures that each point in the target RoI is covered by at least $k$ distinct sensors, where $k$ is a positive integer having a typically value greater than one. A connected $k$-coverage is achieved when each point in the RoI is covered by at least $k$ distinct sensors and each sensor is able to communicate with each other. An optimal $k$-coverage of the network is obtained when each point within the RoI is covered by at least $k$ distinct sensor without any overlapping area. A Coverage Graph (CG) is denoted by \textit{CG(N)=(V, E)}, where \textit{V} denotes the number of vertices and \textit{E} denotes the number of edges. Here, each vertex represents a sensor and an edge represents a link between any two sensors if and only if they fall in the coverage of each other. To construct a CG for a circular RoI, we have built sensor-disjoint cycles within the entire RoI. A WSN rendering two-barrier coverage inside a circular RoI is shown in Fig. \ref{fig:cir}a, and its respective CG is depicted in Fig. \ref{fig:cir}b.

\subsection{Barrier and Barrier Path}

A barrier along the boundaries of a circular RoI can be constructed by taking the union of distinct sensor's coverage. Any arbitrary point from where an intruder may enter the RoI is known as the point of intrusion and any possible path that an intruder may follow to reach the target point is called an intrusion path. In order to provide a guaranteed barrier coverage, there must exist at least one barrier for every possible intrusion path. In this way, any intrusion attempt can be identified and prevented in a timely manner. The maximum number of barrier paths $BP_{max}$ that can be constructed for a given intrusion path without any overlapping coverage is given by equation (\ref{eqn:equ2}).

\begin{equation}
\label {eqn:equ2}
BP_{max}= \floor*{\frac{N} {k}}
\end{equation}
where, \textit{N} represents the total number of sensors and $k$ represents the number of sensors required to ensure $k$ barrier coverage for a possible intrusion path.

\section{Simulation Experiment}
\label{sec:sim-exp}

\begin{table}[b!]
 \centering
    \caption{Simulation parameters.}
    \label{tab:simu}
\resizebox{0.4\textwidth}{!}{%
\begin{tabular}{cc}
\hline\noalign{\smallskip}
\multicolumn{1}{c}{\textbf{Parameters}} & \multicolumn{1}{c}{\textbf{Values}} \\
\hline
Simulator                               & NS-2.35                             \\

Radius of the circular region (R)                             & 40 to 127 {$\mathrm{m}$} \\
Number of sensors (N) & 100 to 400                     \\
Sensing range (R\textsubscript{s})                                     & 15 to 40  {$\mathrm{m}$}                           \\
Transmission range ({R\textsubscript{tx}})                     & 30 to 80 {$\mathrm{m}$}                         \\
Mac type                                & IEEE 802.11                         \\
Sensor's deployment type & (a) Gaussian Distribution\\  & (b) Uniform Distribution \\
Sensing model & Binary sensing model
\\
\noalign{\smallskip}\hline
\end{tabular}}
\end{table}

We have obtained the simulations results using network simulator NS-2.35, which is one of the widely used network simulator to obtain the performance metrics of WSNs. Table \ref{tab:simu} shows different network parameters and their values used to get the simulation results for the number of $k$-barrier paths. Here, we have assumed that the any two sensors can communicate with each other if the transmission range of sensors (\textit{{R\textsubscript{tx}}}) is at least twice the sensing range of sensors (R\textsubscript{s}), \textit{i.e.}, \textit {R\textsubscript{tx} $\ge$ 2R\textsubscript{s}}

\section{Machine Learning Model}
\label{sec:mlmodel}
Machine learning algorithms are broadly classified into supervised and unsupervised learning algorithms. In supervised machine learning, we work with labeled data sets. It is mainly used to solve either regression or classification problems. In contrast, unsupervised machine learning algorithms deal with unlabeled data sets and are mainly used to perform clustering and dimension reduction tasks.

In this study, our objective is to assess the potential of fully connected feed-forward ANN to map the number of $k$-barriers using relevant features.  To evaluate the relevancy of the selected features, we have calculated the feature importance score and performed feature sensitivity analysis.

\subsection{Feature Importance}
This study used the area of the RoI, sensing range, transmission range, and the number of sensors as the potential features and $k$-barriers as the predictand. In machine learning, the selection of input features significantly affects its performance \citep{singh2020machine}. Hence, before training the machine learning model, we have evaluated the relative importance of each selected feature on the predictand. In doing so, we opted regression tree ensemble technique \citep{torres2019regression}. We have first trained a regression tree ensemble model by boosting hundred regression trees using the Least Squares gradient Boosting (LSBoost) ensemble aggregation method (\textit{i.e.,} $r$ = 100), each with a learning rate of one (\textit{i.e.,} $\alpha$ = 1), and the classical decision tree (\textit{i.e.,} decision stumps) has been considered as a weak learner. The LBoost algorithm trains one weak learner at a time and also detects its weak points. Based on such weak points, it generates a new weak learner (\textit{l\textsubscript{i}}) and evaluates its  weight (\textit{i.e.,} $w_{i}$). According to (Eq. \ref{eqn:model}), the algorithm  improves the current model ($M_{i}$) by emphasizing on the prior weak learner's ($M_{i-1}$) weak point. After it has been trained, it integrates the weak learner into the existing model, and creates a single strong learner ($M_{r}$, \textit{i.e.,} ensemble of weak learners) iteratively.
\begin{equation}
    M_{i} =  M_{i-1} + \alpha \; . \; w_{i} \; . \;    l_{i}   \qquad \qquad (i = 1, 2, 3, ..., r)
    \label{eqn:model}
\end{equation}

\par
In addition,  we determine the relative feature importance score by evaluating the overall variations in the node risk ($\Delta R$) due to split on each feature, and then normalising it by the total number of branch nodes ($R_{bn}$). Mathematically, it is represented as in (Eq. \ref{eqn:risk});
\begin{equation}
    \Delta R = \frac{R_{p} - (R_{ch1} + R_{ch2})} {R_{bn}}   
    \label{eqn:risk}
\end{equation}
where, $R_{p}$ denotes the node risk of the parent and $R_{ch1}$ \& $R_{ch2}$ denotes the node risk of two children.  The node risk at individual node (R\textsubscript{i}) is mathematically represented as in (Eq. \ref{eqn:indi}); 
\begin{equation}
   R_{i} = P_{i} \; . \;E_{i}
    \label{eqn:indi}
\end{equation}
where $P_{i}$ denotes the probability  of node \textit{i} and $E_{i}$ denotes the node \textit{i} mean square error.


\subsection{Feature Sensitivity}
Estimating the feature importance score only tells us about the relative importance of each feature. However, it does not convey how the features are associated with the predictand, \textit{i.e.,} whether the predictand value increases with feature (positive impact) or decreases with features (negative impact). To evaluate this, we have performed the sensitivity analysis of the features using Partial Dependence Plot (PDP) \citep{singh2021gaussian,friedman2001greedy}. PDP measures the average effect of a single or more feature by marginalising the effect of all other features taken into consideration. 
We considered the combined impact of two features simultaneously from the input feature set (\textit{i.e.,} $\vartheta$) on the predictand by marginalising the impact of the remaining features. 
To do so, a subset \textit{$\vartheta$\textsuperscript{s}} and a  complimentary set (\textit{$\vartheta$\textsuperscript{c}}) of \textit{$\vartheta$\textsuperscript{s}} is extracted from the feature set (\textit{$\vartheta$ = \{k\textsubscript{1}, k\textsubscript{2}, ..., k\textsubscript{n}\}}) where, n represents the total features. Using (Eq. \ref{eqn:pred}), we can compute any prediction on $\vartheta$.
\begin{equation}
   f(\vartheta) = f(\vartheta^{s}, \vartheta^{c})
    \label{eqn:pred}
\end{equation}
The partial dependence of the feature in \textit{$\vartheta$\textsuperscript{s}} can be determined by calculating the expectation (\textit{E\textsubscript{c}}) of Eq. \ref{eqn:pred}.

\begin{equation}
  \begin{aligned}
    f^{s}(\vartheta^{s}) & =  E_{c}[f(\vartheta^{s}, \vartheta^{c})]\\
      & = \int f(\vartheta^{s}, \vartheta^{c}) \; . \; \rho_{c}(\vartheta^{c}) \; . \; d\vartheta^{c}
  \end{aligned}
   \label{eqn:pred1}
\end{equation}
where, $\rho_{c}(\vartheta^{c}$) indicates the marginal probability of $\vartheta^{c}$, which is represented in Eq. \ref{eqn:mar}.
\begin{equation}
  \rho_{c}(\vartheta^{c}) \approx \int p(\vartheta^{s}, \vartheta^{c}) \; . \; d\vartheta^{s}
    \label{eqn:mar}
\end{equation}
Then, the partial dependency of the feature in $\vartheta^{s}$ can be determined by : 
\begin{equation}
  f^{s}(\vartheta^{s}) \approx  \frac{1} {T} \sum _{i=1}^{T} f(\vartheta^{s}, \vartheta_{i}^{c})\\
    \label{eqn:partial}
\end{equation}
 where, T represents the total number of observations. 
 

\subsection{Model Setup}
In this subsection, we have discussed the architecture of the fully connected feed-forward ANN. We have discussed the working of feed-forward ANN, activation function, and the training algorithm.
\subsubsection{Feed Forward ANN}

 \begin{figure}[t]
\centering
\includegraphics[width=.9\textwidth]{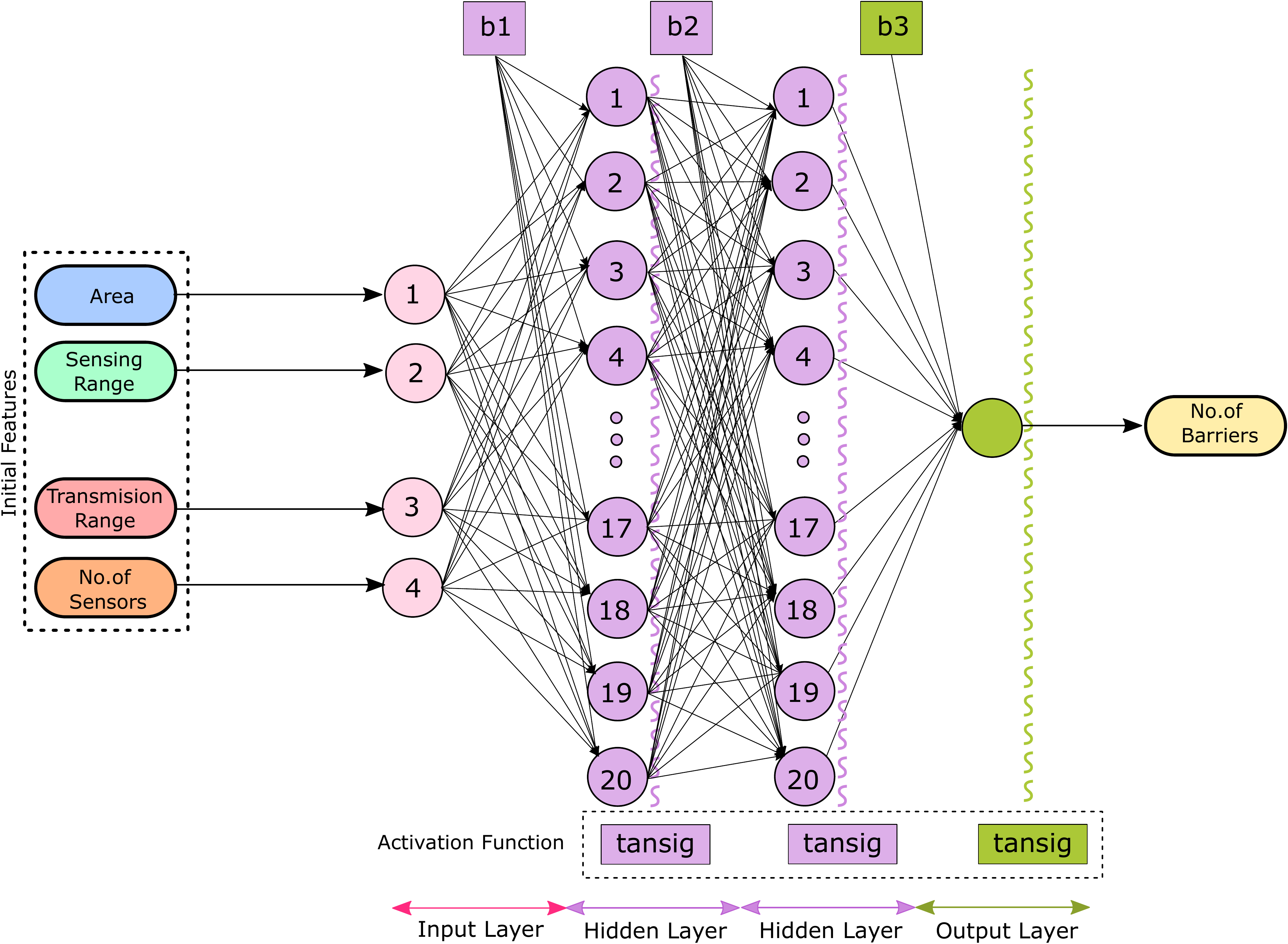}
\caption{Structure of the fully connected feed-forward ANN model having four inputs, two hidden layers having 20 neurons each, and one output (i.e., 4:20:20:1).}
\label{fig:ffnn}
\end{figure}
A Neural Network (NN) is a model which mimics the oversimplification of the brain performance that operates under a particular specific function of interest. The primary objective of the NN model is to discover a mapping function \textit{(f)} which predicts a target function \textit{(f*)} through training the NN using labeled training data sets. During the training phase, the network learns the range of parameters from the training data. Once we trained the model, we need to validate and test the model performance using unseen data. 

A network can be subdivided into basic information-processing elements called neurons, which are the building blocks of ANN.  Layers are groups of neurons, and the network is comprised of interconnections between these layers. There are diverse perspectives of linking layers together that lead to several other forms of NNs like feed-forward neural networks, recurrent neural networks, and convolutional neural networks. In general,  the optimisation problem of a neural network can be represented by Equation (\ref{eq:NN}), where \textit{l} represents the loss function, and \textit{W} the learn-able parameter.  The main objective  is to learn \textit{W} so that the variance between the output of \textit{f} can be minimised, when the input \textit{x\textsubscript{0}} and the actual output \textit{y} are provided. 

\begin{equation}
\label{eq:NN}
    min \; l(f(x_{0},W),y)
\end{equation}

In this study, we have trained a fully connected feed-forward ANN. It is a category of NN formed by organising neurons so that all neurons in each layer are linked to every other neuron in the adjacent forward layer. The data flows mainly in one direction, forward, from the input neurons to the output sensors, passing through hidden sensors (if any). Since each neuron has one activation function, the total activation function for a layer equals the total output value. The training algorithm utilises the output findings to calibrate the sensor connection weight value \citep{novickis2020approach, moldovan2020learning}. Any continuous function can be approximated by a feed-forward ANN with one hidden layer. However, the desired hidden size might be high, making learning unfeasible. Feed-forward ANN is well-suited towards unstructured data, such as data that is not sequential or time-dependent.

In this study, we structured a feed-forward ANN that consists of two hidden layers and one output layer as illustrated in Fig. \ref{fig:ffnn}. Each hidden layers consist of twenty neurons. A common bias value is added to each neuron in the hidden layers, which is followed by an activation function.

\subsubsection{Activation Function}

In feed-forward ANN, the activation function is one of the essential key elements of the neuron \citep{duch1999survey}. It significantly impacts the performance of the neural networks by modifying the neuron output. The Universal Approximation Theorem states that a feed-forward ANN  with one hidden layer and an arbitrary sigmoidal function with adequate sensors can estimate any continuous function with no restrictions on the number of sensors or rather the size of the weights. In this study, we have used the hyperbolic tangent sigmoid transfer function at each layer because it is a bipolar sigmoid function that has a positive response for positive input and a negative response for negative input. Hence, it rid the problem of negative responses for positive values. Further, as the  complexity and non-linear of the problems increases (when we increase the number of sensors and the monitoring area in WSNs), the advantage of using hyperbolic tangent sigmoid transfer function becomes more apparent. The mathematical model is expressed as:


\begin{equation}
    \label{eq:logsig}
    a  = \frac{2}{(1+e^{(-2\cdot n)})-1}
\end{equation}

This expression is mathematically equivalent to tanh(n). However, the computational time complexity of Equation (\ref{eq:logsig}) is lower than tanh(n).

\subsubsection{Training Algorithm}
To minimise the error in the output, the values of weights and biases need to be updated. This is done with the help of a backpropagation training algorithm. In the backpropagation algorithm, the input is transmitted to the hidden layer, which then perpetuates back the sensitivity in order to minimise the error rate by updating the weights and the bias during the process. This algorithm results in low convergence, and in some cases, it also leads to over-fitting. To address these issues, and for quick convergence without over-fitting, approaches like Levenberg-Marquardt backpropagation (LM) and Bayesian Regularization (BR) backpropagation, and Scaled Conjugate Gradient (SCG) backpropagation have been developed. To minimise the sum of squares at every iteration, LM utilises the conjugate gradient backpropagation method. LM is used for curve fitting problems, while SCG is used for pattern recognition problems. Bayesian regularization backpropagation algorithm uses an objective function that incorporates the sum of squared weights and residual sum of squares in order to reduce the prediction errors for attaining the desired model.

In this study, we have used the LM backpropagation algorithm. It is a nonlinear optimisation-based approach for training ANNs, by which it uses second-order derivatives for improved convergence behavior. The LM algorithm offers the features of the steepest descent approach along with the Gauss-Newton method, which provides an invertible matrix, named Hessian matrix \textit{(H)} which is shown in Equation (\ref{eq:Hessian}):

\begin{equation}
    \label{eq:Hessian}
    H(x) \approx J^{T}(x)J(x) + \mu I
\end{equation}
where, $\mu$ represents the combination co-efficient, \textit{J} and \textit{I} represents the Jacobian and identity matrix respectively. The LM modification to the Gauss-Newton algorithm \citep{hagan1994training} is represented in Equation (\ref{eq:gauss}):
\begin{equation}
\label {eq:gauss}
    \Delta x = [J^{T}(x)J(x) + \mu I]^{-1}J^{T}(x)e(x)
\end{equation}
The algorithm seems to be the steepest descent when the value of $\mu$ becomes high, whereas the algorithm is Gauss-Newton when the value of $\mu$ is minimal.  The LM algorithm for the weight update rule  \citep{mathew2018prediction} is determined as a function of Jacobian matrix and error vector (e), which is represented in Equation (\ref{eq:weight}):
\begin{equation}
    \label{eq:weight}
    w(t+1) = w(t) - (J_{t}^{T}J_{t}+ \mu I)^{-1} + J_{t}e_{t}
\end{equation}

\begin{figure}[]
    \centering
    \includegraphics[width=.85\textwidth]{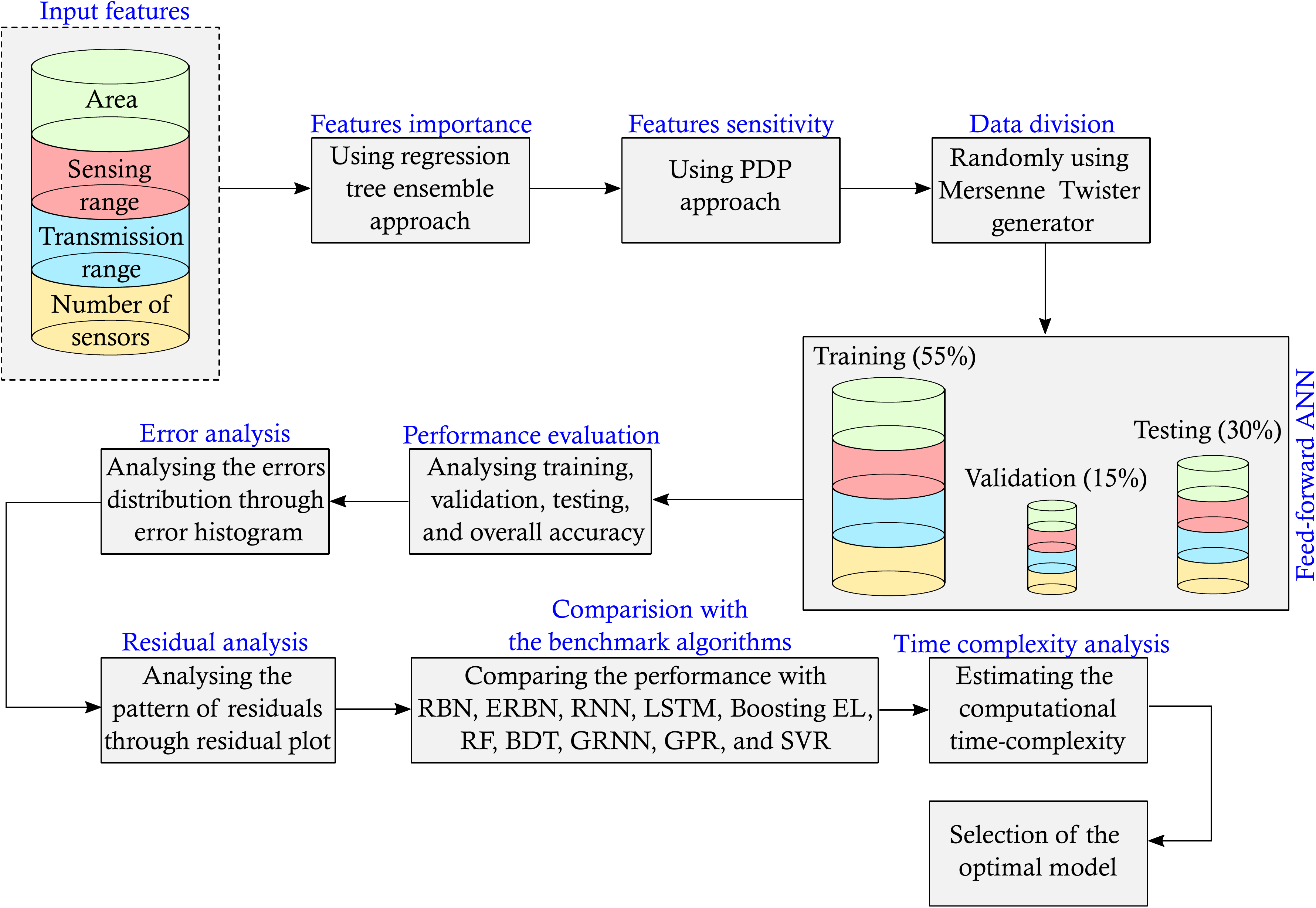}
    \caption{Flowchart of the methodology.}
    \label{fig:flowchart}
\end{figure}

We have randomly divided (using Mersenne Twister generator) the completed data set (182 $\times$ 5) in a 55:15:30 ratio for training, validation, and testing of the feed-forward ANN algorithm. The complete methodology is shown in Fig. \ref{fig:flowchart}.

\newpage 
\begin{figure}[h!]
\centering
\includegraphics[width=.65\textwidth]{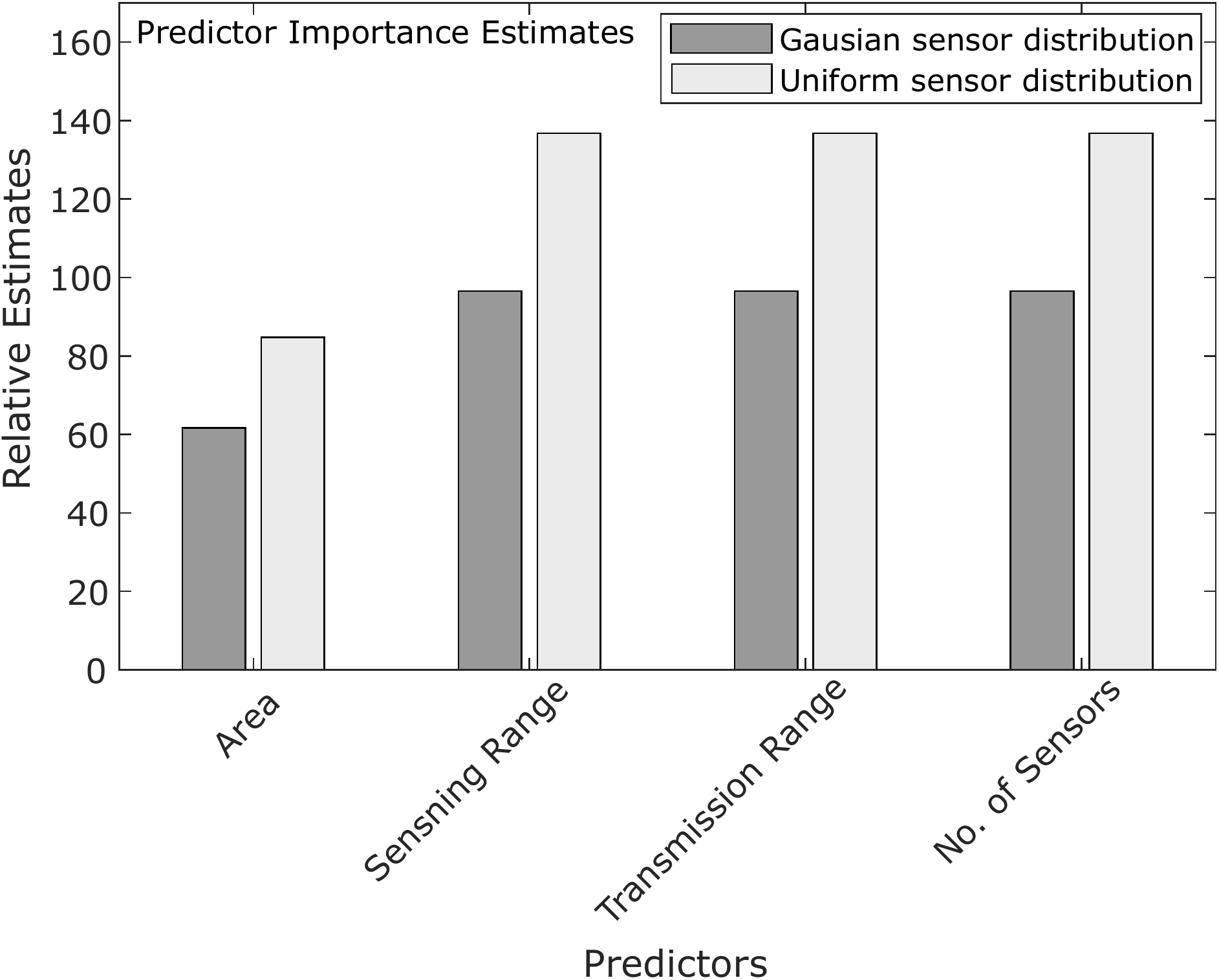}
\caption{Feature importance graph.}
\label{fig:feature_importance}
\end{figure}

\section{Results}
This section presents the results of feature importance analysis, feature sensitivity analysis, and feed-forward ANN.   
\label{sec:results}
\subsection{Feature Importance of Each Features}
Once we evaluated the feature importance of each feature for both Gaussian and uniform sensor distribution, we plotted the relative importance graph for both the scenario (Fig. \ref{fig:feature_importance}). The higher the importance score, the more important is the feature. We found that the relative importance score of the sensing range, transmission range, and sensors is the same and the maximum. In contrast, the area has the least relative importance amongst all the features for both the scenario.

\subsection{Feature Sensitivity Curve}
\begin{landscape}
\begin{figure}[h]
\centering
\includegraphics[width=1.35\textwidth]{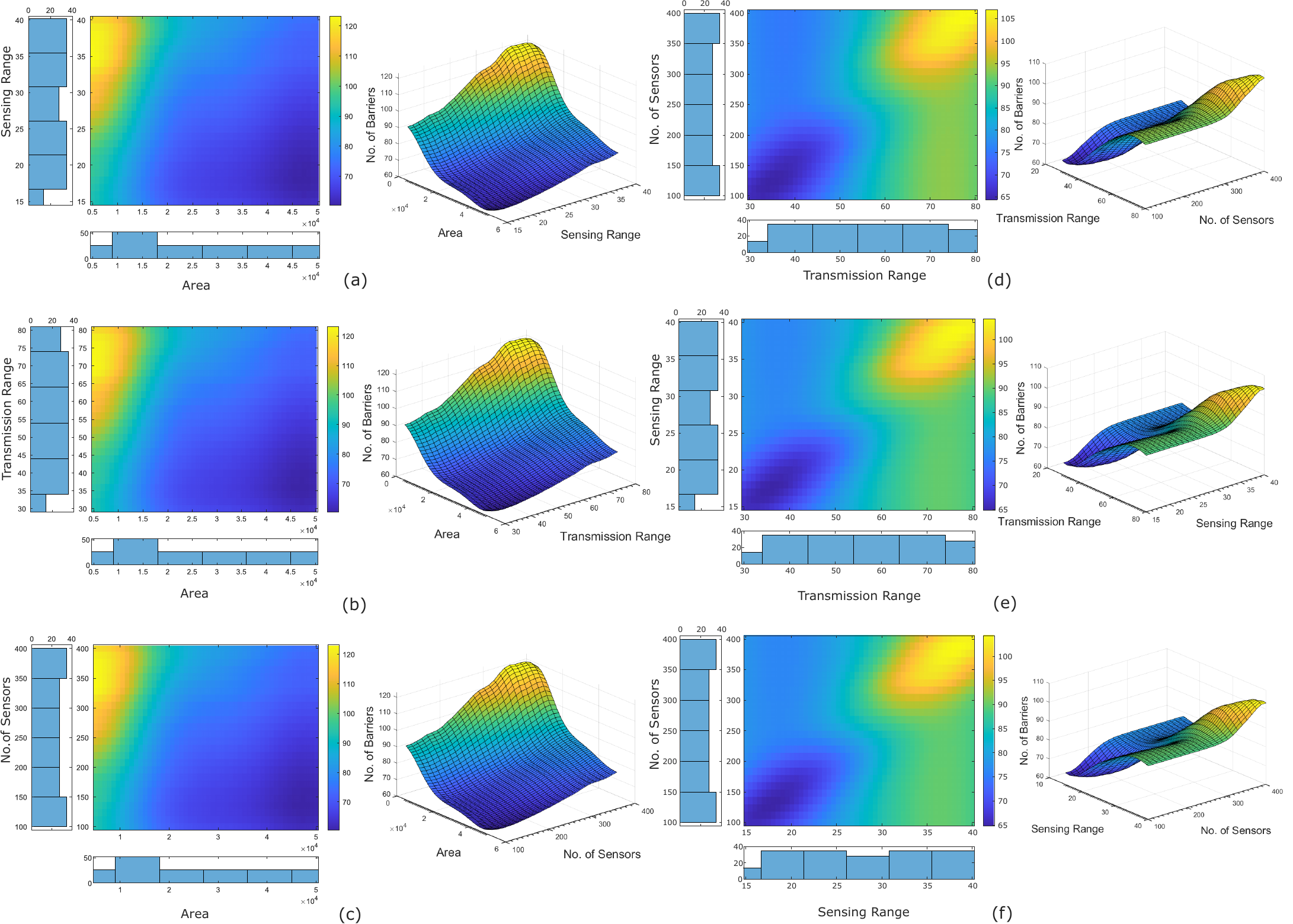}
\caption{Partial dependency plot for circular region considering Gaussian sensor distribution.}
\label{fig:PDP Gaussian_circ}
\end{figure}
\end{landscape}

\begin{landscape}
\begin{figure}[h]
\centering
\includegraphics[width=1.35\textwidth]{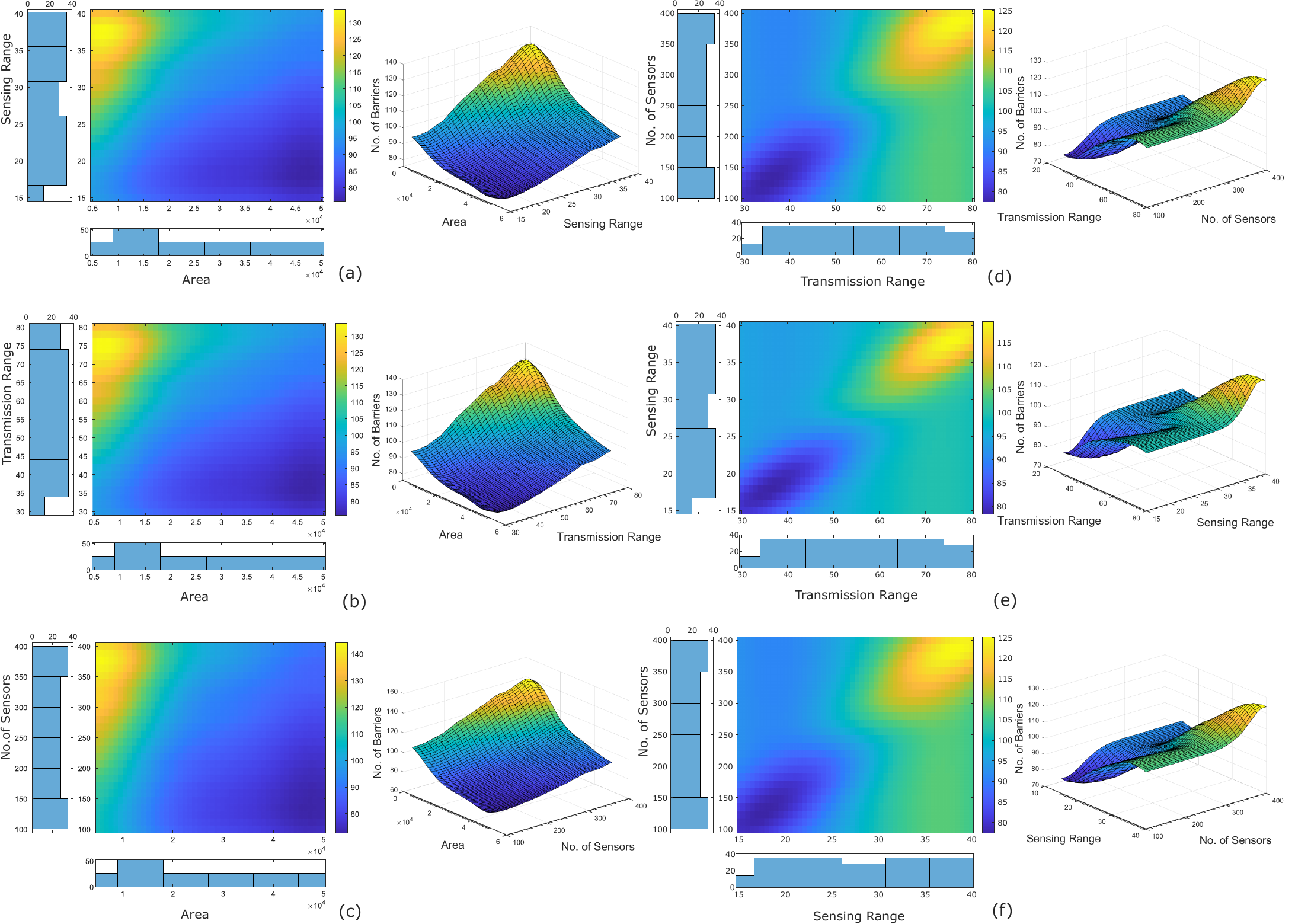}
\caption{Partial dependency plot for circular region considering uniform sensor distribution.}
\label{fig:PDP uniform_circ}
\end{figure}
\end{landscape}

For analysing the feature sensitivity, we have plotted the surface plot of PDP considering two features at a time along with the corresponding two-dimensional plot with an axis-aligned histogram for representing the distribution of the features. For four features, we have six possible sensitivity plots. Again, we have plotted it for both the scenarios \textit{i.e.,} for Gaussian and uniform sensor distribution in Fig. \ref{fig:PDP Gaussian_circ} and \ref{fig:PDP uniform_circ} respectively.

For both scenarios, we observed that the area of the circular region has a negative impact on the number of barriers. In contrast, sensing range, transmission range, and the number of sensors positively impact the number of barriers.  In a nutshell, we observed a similar trend with slight variation in the values for both scenarios.

\subsection{Performance of the Fully Connected Feed-Forward ANN Model}
Once we trained the feed-forward ANN model for Gaussian and Uniform distribution scenarios, we evaluated its performance by plotting a linear regression curve between the predicted and observed barriers. We have used R,  RMSE, and bias as the performance metrics. A high value of R represents that the predicted values are well in accord with the observed value. A low value of RMSE represents a more accurate model. A positive value of bias shows overestimation, and a negative value of bias shows underestimation. Afterward, we have performed error and residual analysis for a robust conclusion.

\begin{figure}[h!]
\centering
\includegraphics[width=1\textwidth]{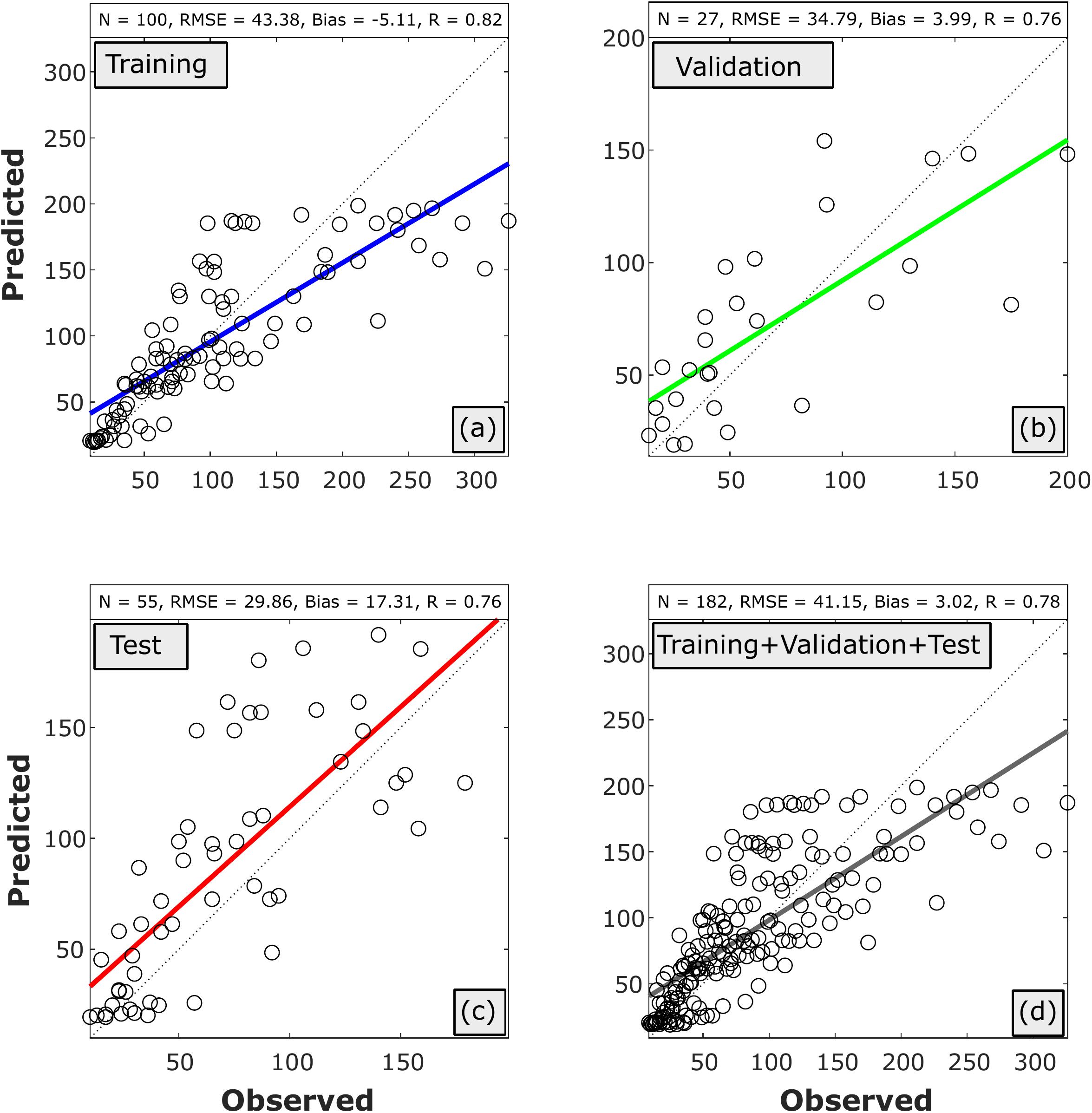}
\caption{Performance of the feed-forward ANN for Gaussian sensor distribution case (a) training accuracy, (b) validation accuracy, (c) testing accuracy, and (d) overall accuracy.}
\label{fig:circ_g_result}
\end{figure}

\subsubsection{For Gaussian Sensor Distribution}
To evaluate the performance of the trained feed-forward ANN for Gaussian sensor distribution, we have reported the training, validation, testing, and overall accuracy. For training accuracy, we feed the training data set as input to the trained feed-forward ANN and evaluated its performance. We found that the trained model work reasonably well on the training data set with R = 0.82, RMSE = 43.38, and bias = -5.11 (Fig. \ref{fig:circ_g_result}a). The presence of a small negative bias indicates that the values are slightly getting underestimated. Evaluating the model performance using the test data results in bias study and hence its performance needs to be evaluated using the unseen/new data sets. In doing so, we have first validated the trained feed-forward ANN model through validation data set (Fig. \ref{fig:circ_g_result}b). We found that the trained model performs well and results in a good fit (with R = 0.76, RMSE = 34.79, and bias = 3.99) while tuning the hyper-parameters. Afterward, we used the test data for unbiased evaluation of the trained model (Fig. \ref{fig:circ_g_result}c). We observed that the trained model perform well on the test data (with R = 0.76, RMSE = 29.86, and bias = 17.31). The predicted value accord well with the observed value with slight scattering. The presence of positive bias represents that the values are slightly overestimated during the testing phase. Finally, we combined all the data sets together (training, validation, and testing) and fed it into the trained feed-forward ANN model to calculate the model's overall accuracy. We found that the trained model perform reasonably well on the complete data sets with R = 0.78, RMSE = 41.15, and bias = 3.02 (Fig. \ref{fig:circ_g_result}d).

\begin{figure}[h!]
\centering
\includegraphics[width=.8\textwidth]{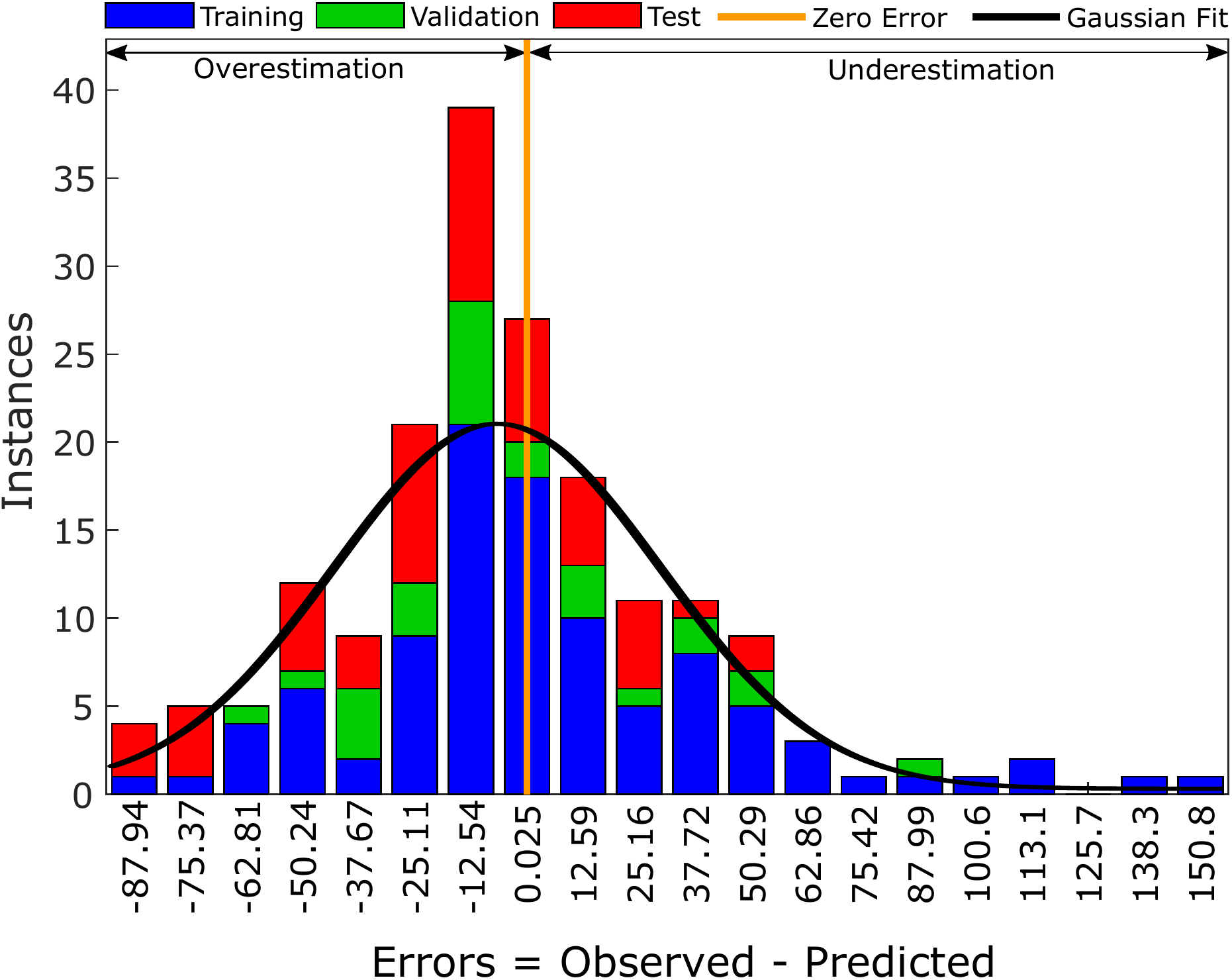}
\caption{Error analysis with error histogram with 20 bin size for Gaussian sensor distribution case.}
\label{fig:circ_g_errorhist}
\end{figure}

Further, to analyse the error distribution during the training, validation, and testing phase, we have performed error analysis and plotted the combined error histogram using twenty bin size (Fig. \ref{fig:circ_g_errorhist}). The combined error from the trained feed-forward ANN model ranges from -87.94 (leftmost bin) to 150.8 (rightmost bin) and follows a slightly right-skewed Gaussian distribution. The peak of the distribution lies near the zero error line indicating a more accurate model. The region left to the zero error line indicates overestimated region, and the one on the right represents an underestimated region. Overall, the number of instances in the overestimated region is higher than the underestimated region results in the overestimation of predicted values by the trained deep learning model. This statement is validated by the presence of a positive bias of 3.02 (Fig. \ref{fig:circ_g_result}d).

\begin{figure}[h!]
\centering
\includegraphics[width=1\textwidth]{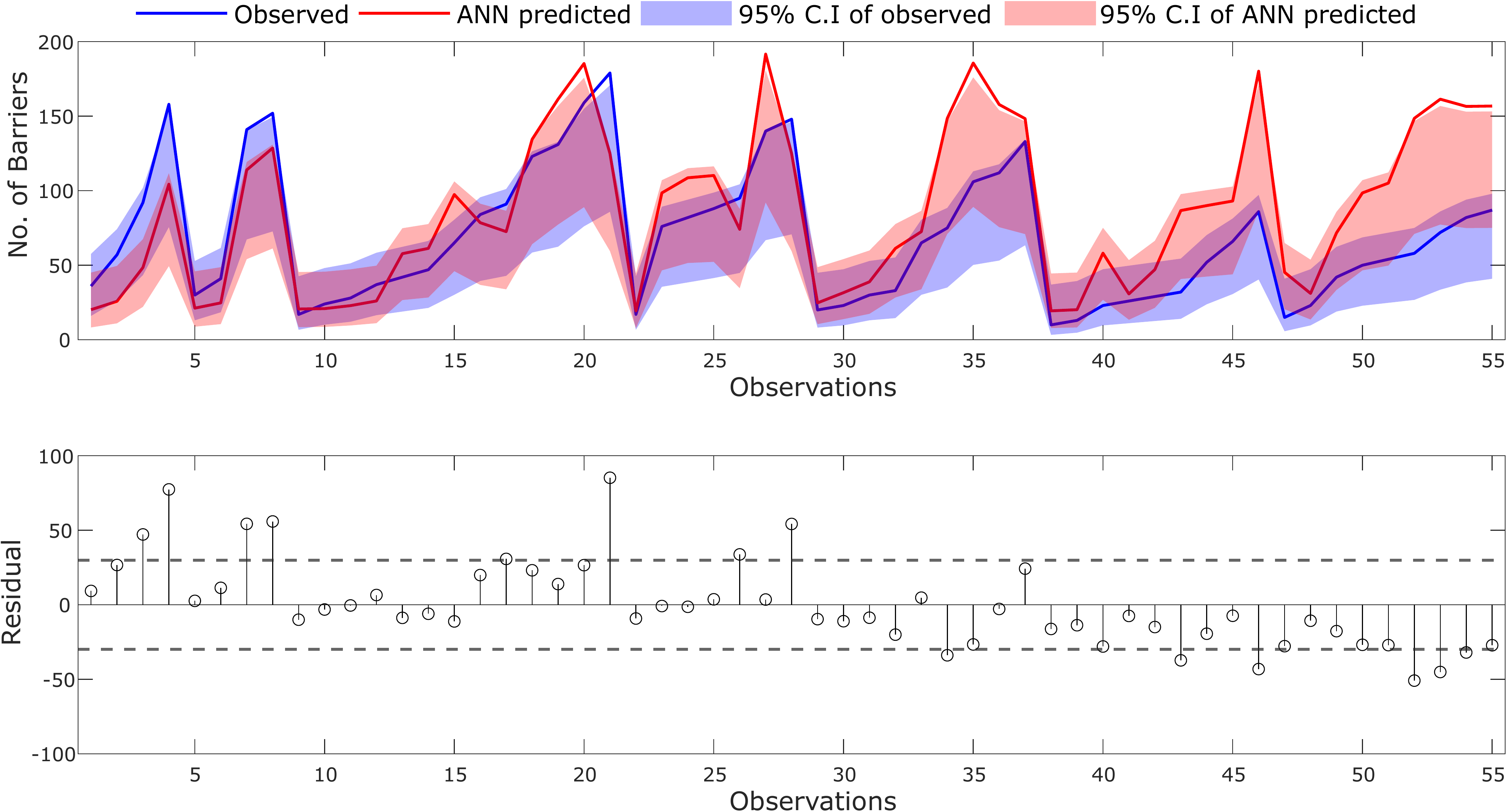}
\caption{Residual analysis of the feed-forward ANN output for Gaussian sensor distribution case.}
\label{fig:circ_g_residual}
\end{figure}

Furthermore, to evaluate the appropriateness of the trained model, we have performed the residual analysis and plotted the time series plot of the test data along with the corresponding residual plot. We have plotted the observed values (in blue) and predicted values (in red) along with their 95\% Confidence Interval (C.I). The dashed line represents the RMSE value of the testing phase. The residuals are well scattered and do not follows any specific pattern indicating a good fit (Fig. \ref{fig:circ_g_residual}).

\subsubsection{For Uniform Sensor Distribution}

\begin{figure}[h!]
\centering
\includegraphics[width=1\textwidth]{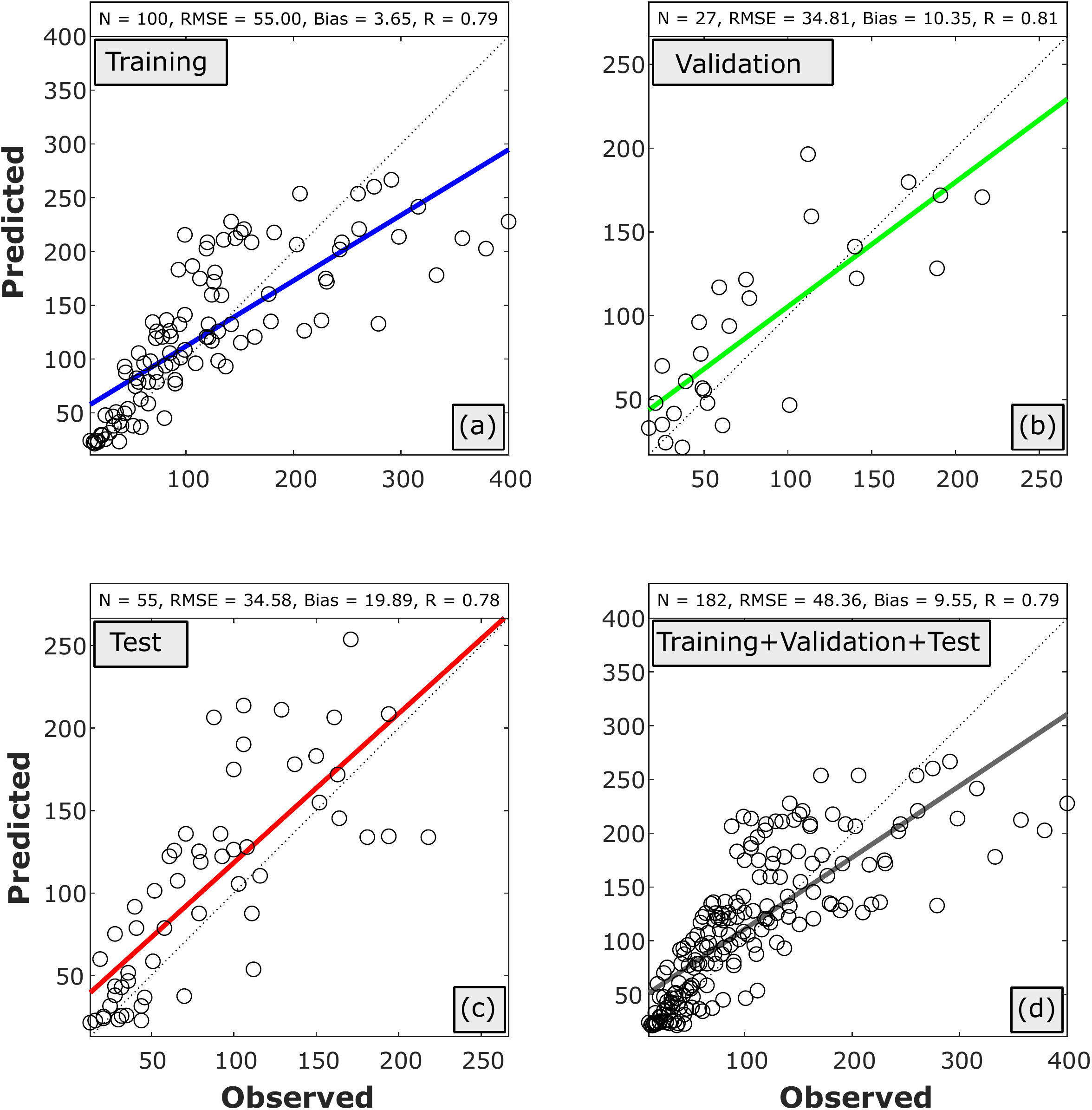}
\caption{Performance of the feed-forward ANN for uniform sensor distribution case (a) training accuracy, (b) validation accuracy, (c) testing accuracy, and (d) overall accuracy.}
\label{fig:circ_u_result}
\end{figure}

Similar to the Gaussian sensor distribution scenario, we have also evaluated the performance of the deep learning model for uniform sensor distribution. We reported the training, validation, testing, and overall accuracy of the feed-forward ANN model that we trained for uniform sensor distribution. For training accuracy, we have evaluated the model over the training data sets. In doing so, we observed that the model performs quite well over the training data sets (Fig. \ref{fig:circ_u_result}a). The predicted values are close to observed values (with R = 0.79, RMSE = 55, and bias = 3.65). However, the RMSE is high, and R is slightly low as compared with the training accuracy of the Gaussian sensor distribution. Afterward, we feed the validation data sets to the model input to report the validation accuracy. The predicted values (while tuning the hyper-parameters) are in agreement with the observed values with R = 0.81, RMSE = 34.81, and bias = 10.35 (Fig. \ref{fig:circ_u_result}b). For unbiased evaluation, we have evaluated the trained model performance over the test data set. We observed that the trained model performs well over the test data with  R = 0.78, RMSE = 34.58, and bias = 19.89 (Fig. \ref{fig:circ_u_result}c). Finally, we feed the combined data into the model to report the overall accuracy. 

\begin{figure}[t]
\centering
\includegraphics[width=.8\textwidth]{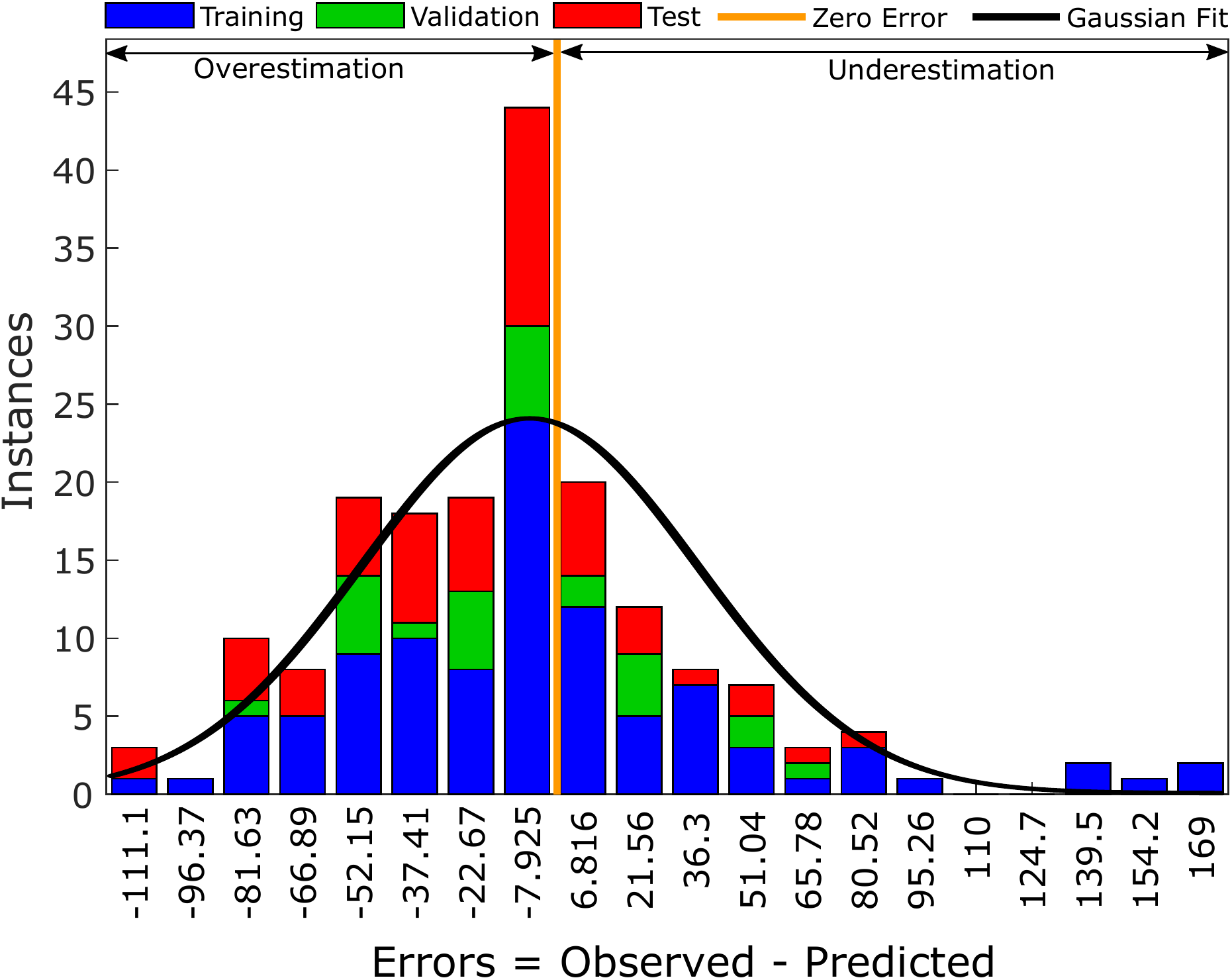}
\caption{Error analysis with error histogram with 20 bin size for uniform sensor distribution case.}
\label{fig:circ_u_errorhist}
\end{figure}
\begin{figure}[h!]
\centering
\includegraphics[width=1\textwidth]{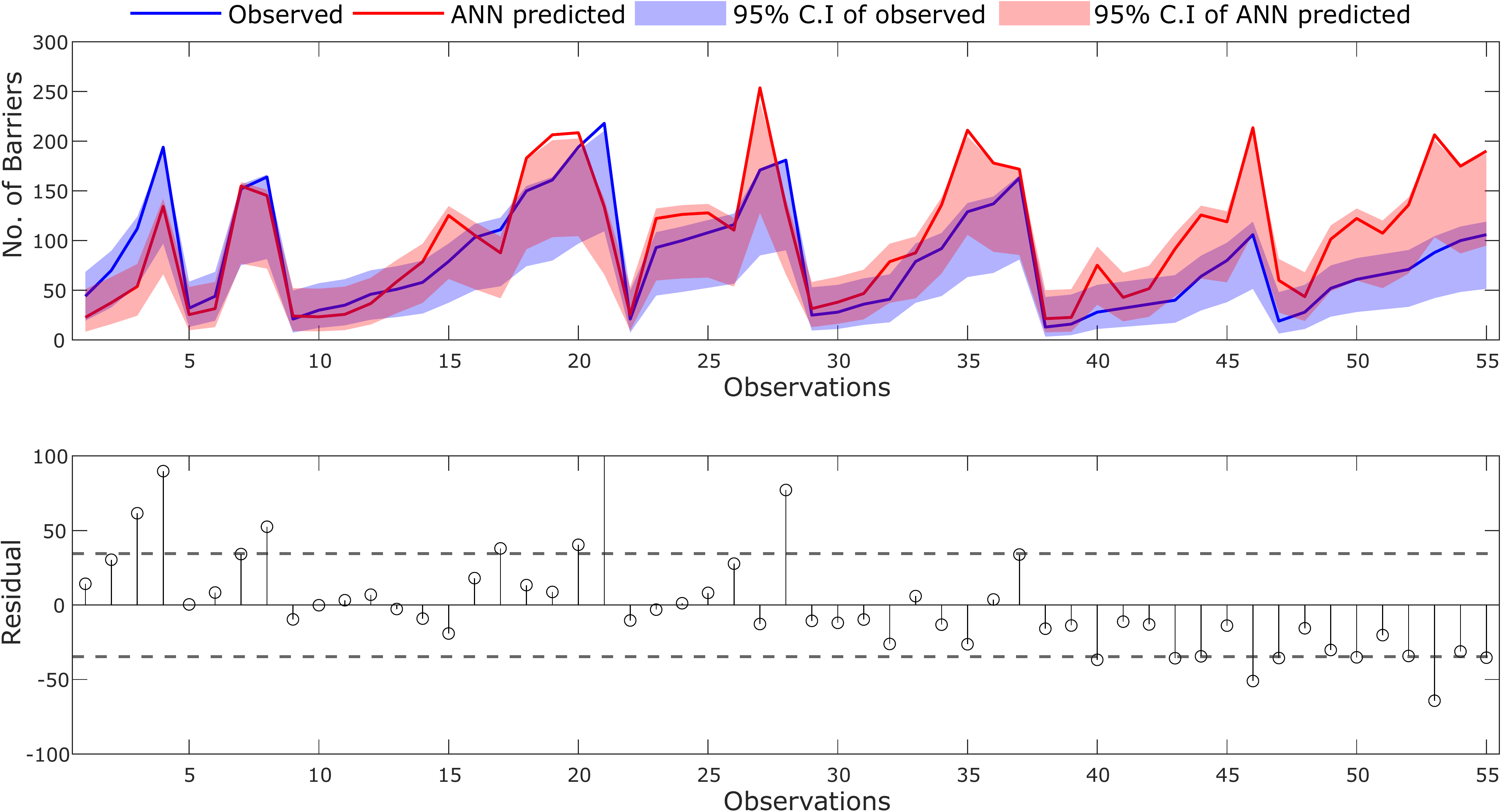}
\caption{Residual analysis of the feed-forward ANN output for uniform sensor distribution case.}
\label{fig:circ_u_residual}
\end{figure}
We found that the trained model also performs well over the complete data set with R = 0.79, RMSE = 48.36, and bias 9.55 (Fig. \ref{fig:circ_u_result}d). 

Further, we perform the error analysis to understand the distribution of error in a uniform sensor distribution scenario. In doing so, we plotted the combined error histogram using twenty bin size (Fig. \ref{fig:circ_u_errorhist}). We observed a similar trend as for the Gaussian sensor distribution scenario despite the high error range. The total error ranges from -111.1 (leftmost bin) to 169 (rightmost bin) and follows a slightly right-skewed Gaussian distribution. Here also, the peak of the distribution lies near the zero error line indicating an accurate model. The total number of instances in the overestimated region is higher than the underestimated region. Due to this, a positive bias is present in the model.

Furthermore, we have performed the residual analysis and plotted the time series plot of observed-predicted values for the testing phase along with the corresponding residual plot (Fig. \ref{fig:circ_u_residual}). Here, the residuals are well scattered and do not follow any particular path or pattern, indicating that the linear plot is a good fit.

\section{Discussion}
\label{sec:discussion}
This study uses fully connected feed-forward ANN to predict the number of $k$-barriers for intrusion detection in WSNs. We trained two separate feed-forward ANN models for Gaussian and uniform sensor distribution scenarios. We observed that the proposed architecture of a fully connected feed-forward ANN model gives promising results for both scenarios. Although the correlation coefficient value is nearly equal for both scenarios, the RMSE and bias value for the Gaussian sensor distribution scenario is better. 

\subsection{Comparing with different variant of feed-forward ANN}
For an unbiased evaluation, we have generated different scenarios of the feed-forward ANN model based on the number of hidden layers used, ranging from shallow to deep feed-forward ANN model (Table \ref{table:layervariation}). To do so, we have selected six different scenarios corresponding to 1, 2, 3, 4, 5, and 10 hidden layers. A feed-forward ANN model with more than ten hidden layers will result in high time complexity and eventually not an optimal solution for intrusion detection, which is a time-sensitive application.  We have reported the training, validation, testing, and overall accuracy for all six scenarios. Based on the overall performance, we have categorised the performance of each scenario into poor, fair, and satisfactory.
We found that the feed-forward ANN with 2, 3, and 10 layers resulted in satisfactory performance. Out of these three scenarios, feed-forward ANN with two layers (\textit{i.e.,} 4:20:20:1) shows the best performance.
\begin{table}[]
\caption{Comparison of the performance of different scenarios of feed-forward ANN.}
\resizebox{\textwidth}{!}{  
\begin{tabular}{ccccccccccccccc}
\hline
\multicolumn{2}{c}{\multirow{2}{*}{\textbf{Scenarios}}} & \multicolumn{3}{c}{\textbf{Training}} & \multicolumn{3}{c}{\textbf{Validation}} & \multicolumn{3}{c}{\textbf{Testing}} & \multicolumn{3}{c}{\textbf{Overall}} & \multirow{2}{*}{\textbf{Performance}} \\ \cline{3-14}
\multicolumn{2}{c}{} & \textbf{R} & \textbf{RMSE} & \textbf{Bias} & \textbf{R} & \textbf{RMSE} & \textbf{Bias} & \textbf{R} & \textbf{RMSE} & \textbf{Bias} & \textbf{R} & \textbf{RMSE} & \textbf{Bias} &  \\ \hline
\multirow{2}{*}{\textbf{\begin{tabular}[c]{@{}c@{}}4:20:1\\ (Single layer)\end{tabular}}} & \textbf{Gaussian} & 0.40 & 60.66 & -11.69 & 0.28 & 53.27 & -17.26 & 0.55 & 61.46 & 19.05 & 0.44 & 59.69 & -3.23 & Fair \\ \cline{2-15} 
 & \textbf{Uniform} & 0.51 & 67.08 & 6.44 & 0.53 & 52.25 & 8.81 & 0.53 & 52.25 & 8.81 & 0.56 & 65.06 & 17.31 & Fair \\ \hline
\multirow{2}{*}{\textbf{\begin{tabular}[c]{@{}c@{}}4:20:20:1\\ (Two layers)\end{tabular}}} & \textbf{Gaussian} & 0.82 & 43.38 & -5.11 & 0.76 & 34.79 & 3.99 & 0.76 & 29.86 & 17.31 & 0.78 & 41.15 & 3.02 & Satisfactory \\ \cline{2-15} 
 & \textbf{Uniform} & 0.79 & 55.00 & 3.65 & 0.81 & 34.81 & 10.35 & 0.78 & 34.58 & 19.89 & 0.79 & 48.36 & 9.55 & Satisfactory \\ \hline
\multirow{2}{*}{\textbf{\begin{tabular}[c]{@{}c@{}}4:20:20:20:1\\ (Three layers)\end{tabular}}} & \textbf{Gaussian} & 0.81 & 42.94 & -2.91 & 0.75 & 44.94 & 13.66 & 0.72 & 37.51 & 2.08 & 0.76 & 42.93 & 1.06 & Satisfactory \\ \cline{2-15} 
 & \textbf{Uniform} & 0.82 & 49.54 & 14.69 & 0.80 & 48.62 & 18.31 & 0.66 & 47.90 & 22.24 & 0.77 & 50.17 & 17.51 & Satisfactory \\ \hline
\multirow{2}{*}{\textbf{\begin{tabular}[c]{@{}c@{}}4:20:20:20:20:1\\ (Four layers)\end{tabular}}} & \textbf{Gaussian} & 0.44 & 59.93 & -10.18 & 0.30 & 58.59 & -20.99 & 0.54 & 57.61 & -12.55 & 0.47 & 58.74 & -12.50 & Fair \\ \cline{2-15} 
 & \textbf{Uniform} & 0.08 & 78.32 & -55.49 & 0.01 & 70.94 & -43.85 & 0.12 & 81.12 & -84.73 & 0.09 & 78.08 & -62.60 & Poor \\ \hline
\multirow{2}{*}{\textbf{\begin{tabular}[c]{@{}c@{}}4:20:20:20:20:20:1\\ (Five layers)\end{tabular}}} & \textbf{Gaussian} & 0.46 & 62.44 & -24.53 & 0.28 & 67.11 & -17.13 & 0.42 & 50.32 & -11.17 & 0.43 & 60.00 & -19.40 & Fair \\ \cline{2-15} 
 & \textbf{Uniform} & 0.11 & 83.11 & -102.25 & 0.06 & 84.13 & -97.75 & 0.08 & 62.41 & -70.25 & 0.07 & 78.22 & -91.91 & Poor \\ \hline
\multirow{2}{*}{\textbf{\begin{tabular}[c]{@{}c@{}}4:20:20: ... :20:20:1\\ (Ten layers)\end{tabular}}} & \textbf{Gaussian} & 0.64 & 48.35 & 25.09 & 0.82 & 46.20 & 9.78 & 0.75 & 45.01 & 22.25 & 0.69 & 47.72 & 21.96 & Satisfactory \\ \cline{2-15} 
 & \textbf{Uniform} & 0.75 & 47.90 & 0.44 & 0.78 & 59.69 & -3.47 & 0.70 & 59.17 & 5.20 & 0.73 & 53.23 & 1.30 & Satisfactory \\ \hline
\end{tabular}
}
\label{table:layervariation}
\end{table}

\begin{table}[b]
\caption{Comparison with the benchmark algorithms.}
\resizebox{\textwidth}{!}{  
\begin{tabular}{ccccccccccccccccccccccc}
\hline
\multirow{3}{*}{\centering \rotatebox{90}{\textbf{\begin{tabular}[c]{@{}c@{}}Performance\\  metrics \end{tabular} }}} & \multicolumn{22}{c}{\textbf{Methods}} \\ \cline{2-23} 
 & \multicolumn{2}{c}{\textbf{Feed forward ANN}} & \multicolumn{2}{c}{\textbf{RBN}} & \multicolumn{2}{c}{\textbf{ERBN}} & \multicolumn{2}{c}{\textbf{RNN}} & \multicolumn{2}{c}{\textbf{LSTM}} & \multicolumn{2}{c}{\textbf{Boosting EL}} & \multicolumn{2}{c}{\textbf{Bagging EL (RF)}} & \multicolumn{2}{c}{\textbf{BDT}} & \multicolumn{2}{c}{\textbf{GRNN}} & \multicolumn{2}{c}{\textbf{GPR}} & \multicolumn{2}{c}{\textbf{SVR}} \\ \cline{2-23} 
 & \rotatebox{90}{\textbf{Gaussian }} & \rotatebox{90}{\textbf{Uniform}} & \rotatebox{90}{\textbf{Gaussian}} & \rotatebox{90}{\textbf{Uniform}} & \rotatebox{90}{\textbf{Gaussian}} & \rotatebox{90}{\textbf{Uniform}} & \rotatebox{90}{\textbf{Gaussian}} & \rotatebox{90}{\textbf{Uniform}} & \rotatebox{90}{\textbf{Gaussian}} & \rotatebox{90}{\textbf{Uniform}} & \rotatebox{90}{\textbf{Gaussian}} & \rotatebox{90}{\textbf{Uniform}} & \rotatebox{90}{\textbf{Gaussian}} & \rotatebox{90}{\textbf{Uniform}} & \rotatebox{90}{\textbf{Gaussian}} & \rotatebox{90}{\textbf{Uniform}} & \rotatebox{90}{\textbf{Gaussian}} & \rotatebox{90}{\textbf{Uniform}} & \rotatebox{90}{\textbf{Gaussian}} & \rotatebox{90}{\textbf{Uniform}} & \rotatebox{90}{\textbf{Gaussian}} & \rotatebox{90}{\textbf{Uniform}} \\ \hline
\textbf{R} & 0.78 & 0.79 & 0.68 & 0.34 & 0.69 & 0.68 & 0.97 & 0.96 & 0.08 & 0.07 & 0.90 & 0.89 & 0.99 & 0.99 & 0.93 & 0.93 & 0.89 & 0.88 & 0.98 & 0.98 & 0.63 & 0.66 \\ \hline
\textbf{RMSE} & 41.15 & 48.36 & 65.67 & 168.51 & 75.12 & 89.68 & 15.71 & 22.27 & 1.16 & 1.42 & 37.31 & 46.00 & 11.37 & 12.63 & 29.29 & 32.90 & 39.66 & 48.80 & 11.49 & 13.72 & 48.44 & 52.22 \\ \hline
\textbf{Bias} & 3.02 & 9.55 & 39.19 & 137.51 & 60.11 & 71.11 & 61.45 & 65.36 & -17.10 & -20.9 & 59.06 & 69.96 & 53.90 & 62.35 & 57.42 & 63.22 & 60.70 & 71.69 & 53.18 & 61.07 & 18.24 & 20.26 \\ \hline
\end{tabular}
}
\label{table:benchmark}
\end{table}

\subsection{Comparison with the benchmark algorithms.}
Various other findings have been reported for high intrusion detection accuracy through the machine learning approach \citep{safaldin2021improved,nancy2020intrusion}.  Hence, any conclusion based on comparing different scenarios of a single algorithm may result in a biased conclusion. To ensure a fair evaluation of the proposed approach, we have compared the results of feed-forward ANN with the other benchmark algorithms using R, RMSE, and bias as the performance metrics. We have selected Radial Basis Neural Network (RBN), Exact Radial Basis Neural Network (ERBN), Recurrent Neural Network (RNN), LSTM, Boosting (Least-square boosting) Ensemble Learning (EL), Bagging EL (Random Forest), Binary Decision Tree (BDT), General Regression Neural Network (GRNN), Gaussian Process Regression (GPR), and Support Vector Regression (SVR) as potential benchmark algorithms because these algorithms are amongst best performing algorithms in the black-box and explainable based machine learning category \citep{lundberg2018explainable,zhang2020hybrid,elias2020genetic,lin2020radial,mansor2020systematic,singh2021machine,roscher2020explainable,belle2021principles}. On comparing, we found that the feed-forward ANN outperforms all the benchmark algorithms in terms of R, RMSE and bias (Table \ref{table:benchmark}). Other than the feed-forward ANN, SVR performs well and ranks second among the benchmark algorithms. Interestingly, we found that some benchmark algorithms show a strong correlation (\textit{i.e.,} high R value); however, they produce biased results. The bias values are very high, indicating that these models strongly overestimate the response variable.

\begin{figure}[t]
    \centering
    \includegraphics[width=\textwidth]{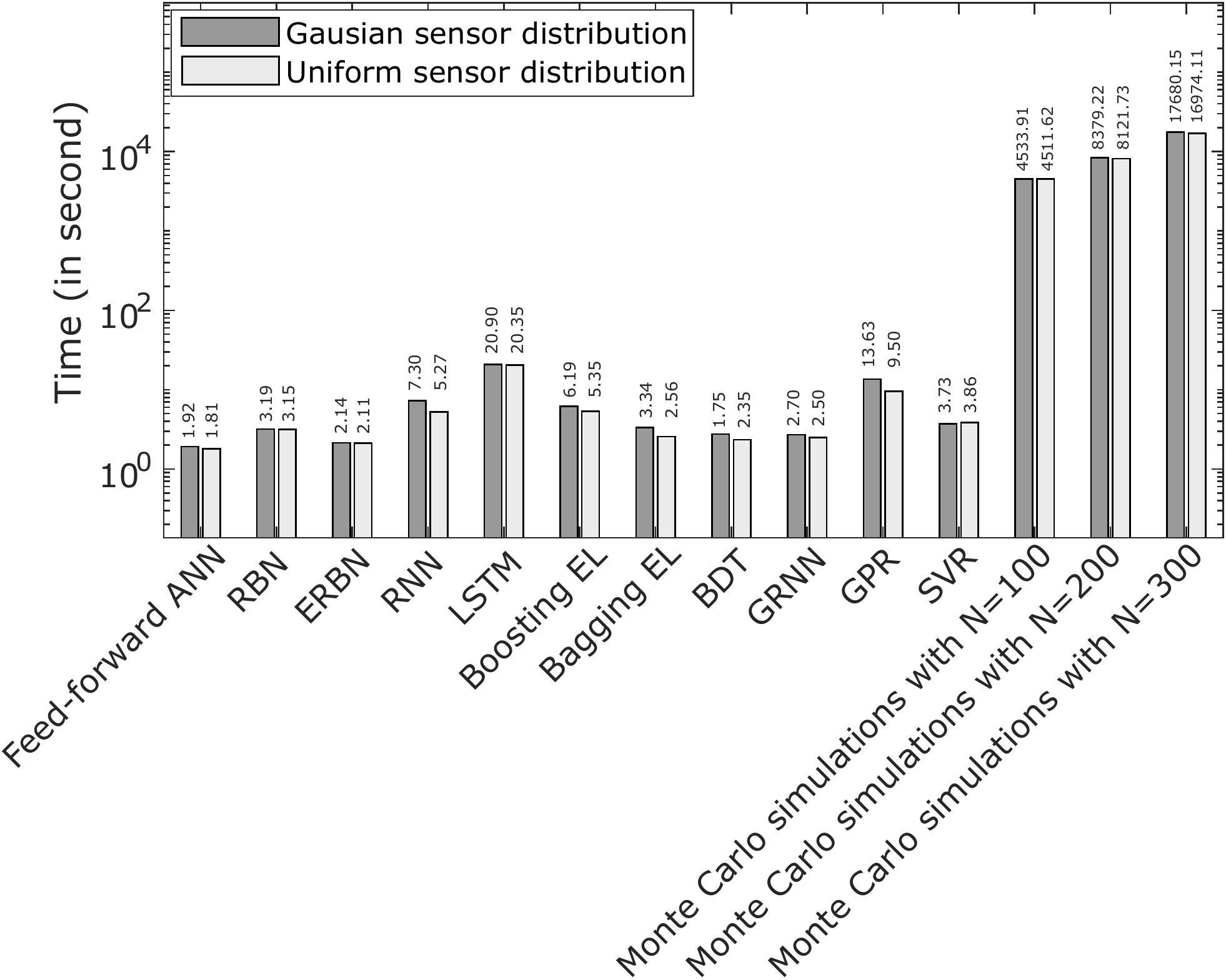}
    \caption{Comparison of the computational time complexity of the feed-forward ANN with the benchmark algorithms with three different scenarios of Monte Carlo simulations. A log scale is used for the vertical axis.}
    \label{fig:time_complexity}
\end{figure}

\subsection{Comparison of the computational time complexity}
Further, we have also compared the performance of the algorithms as mentioned above in terms of computational efficiency. We have estimated and plotted the computational time complexity graph of all the three algorithms for both scenarios (Fig. \ref{fig:time_complexity}). We observed that the feed-forward ANN algorithm exhibits the lowest, and the LSTM exhibits the highest computational time complexity. Apart from this, we have also plotted the time complexity of the Monte Carlo simulation for three different scenarios. We have estimated the time-complexity for sensors as 100, 200, and 300. We have kept the other features constant (area $\approx$ 5000, sensing range = 15, and transmission range = 30).  We observed that the computation time-complexity of the Monte Carlo simulation increases with the number of sensors. This shows the efficacy and need of the proposed deep learning architecture to cut down the computational cost during the network setup time.

\section{Conclusion}
\label{sec:conclusion}
This study presented a fully connected feed-forward ANN architecture for the accurate mapping of the number of $k$-barriers for intrusion detection using WSNs. In doing so, we have trained two separate feed-forward ANN models for both Gaussian and uniform sensor distribution using four potential features extracted through simulations. While evaluating the feature importance and feature sensitivity for Gaussian and uniform sensor distribution scenario, we observed that the sensing range, transmission range, and the number of sensors are the most relevant features amongst all which positively impact the predictand. In contrast, the area is the least important feature, and it is negatively related to the predictand. After training the models, we have evaluated their performance over the unseen data and found that both models give promising results. 

Further, we have compared the performance of the feed-forward ANN with the benchmark algorithms. We found that the proposed feed-forward architecture outperforms the benchmark algorithms in terms of accuracy and computational time complexity. 

This study is a step toward the accurate and time-efficient prediction of the number of $k$-barriers for intrusion detection using WSNs. Our methodology can be used to cut down the time and cost requirement during practical network setup.

\section*{CRediT author statement}
\noindent \textbf{ Abhilash Singh:} Conceptualisation, Methodology, Software, Data Curation, Validation, Writing- Original draft preparation, Visualization, Investigation, Writing- Reviewing and Editing.\\
 \textbf{ J. Amutha:} Conceptualisation, Methodology, Software, Data Curation, Validation, Writing- Original draft preparation, Visualization, Writing- Reviewing and Editing.\\
\textbf{ Jaiprakash Nagar:} Conceptualisation, Methodology, Data Curation, Visualization, Writing- Original draft preparation, Visualization, Writing- Reviewing and Editing.\\
\textbf{ Sandeep Sharma:}  Methodology, Data Curation, Visualization, Investigation, Writing- Reviewing and Editing, Supervision, Project Administration.\\

\section*{Data and code availability}
\noindent The datasets generated during and/or analysed during the current study can be downloaded from \href{https://www.kaggle.com/datasets/abhilashdata/ffannid-intrusion-detection-in-wsns}{here (Data)} and the code can be downloaded from \href{https://in.mathworks.com/matlabcentral/fileexchange/116670-a-deep-learning-approach-to-predict-the-number-of-k-barriers?s_tid=prof_contriblnk}{here (Code)} (accessed on 25 August 2022). 

\section*{Acknowledgments}
\noindent The authors would like to acknowledge IISER Bhopal, MITS Gwalior and IIT Kharagpur for providing institutional support. They would like to thank to the editor and all the anonymous reviewers for providing helpful comments and suggestions.

\bibliography{mybibfile}

\end{document}